\documentclass[11pt]{article}
\usepackage[font=small,labelfont=bf]{caption}
\usepackage[numbers,sort&compress]{natbib}
\usepackage{siunitx}
\usepackage{tabularx}
\usepackage{booktabs}
\usepackage{caption}
\usepackage{bm}

\usepackage{lineno,hyperref}
\usepackage{bbm}
\usepackage{float}
\usepackage{bbding}
\usepackage{ulem}
\usepackage{algpseudocode}
\usepackage[skip=0pt]{caption}
\usepackage{algorithm}
\usepackage[export]{adjustbox}
\usepackage{multicol,tabularx,capt-of}
\usepackage{hhline}
\usepackage{multirow}
\usepackage[utf8]{inputenc}
\usepackage[british]{babel}
\usepackage{tabularx,ragged2e,booktabs,caption}
\graphicspath{ {./figures/} }	
\newcommand{\blind}{0}
	
\makeatletter
\renewcommand\section{\@startsection {section}{1}{\z@}%
                                       {-3.5ex \@plus -1ex \@minus -.2ex}%
                                       {2.3ex \@plus.2ex}%
                                       {\normalfont\fontfamily{phv}\fontsize{16}{19}\bfseries}}
\renewcommand\subsection{\@startsection{subsection}{2}{\z@}%
                                         {-3.25ex\@plus -1ex \@minus -.2ex}%
                                         {1.5ex \@plus .2ex}%
                                         {\normalfont\fontfamily{phv}\fontsize{14}{17}\bfseries}}
\renewcommand\subsubsection{\@startsection{subsubsection}{3}{\z@}%
                                        {-3.25ex\@plus -1ex \@minus -.2ex}%
                                         {1.5ex \@plus .2ex}%
{\normalfont\normalsize\fontfamily{phv}\fontsize{14}{17}\selectfont}}
\makeatother
	
\usepackage{amsmath, amsfonts, amssymb}
\usepackage{graphicx}
\usepackage{enumerate}
\usepackage{xcolor}
\usepackage{natbib} 
\usepackage{url} 
	
\begin{document}
		
\def\spacingset#1{\renewcommand{\baselinestretch}%
			{#1}\small\normalsize} \spacingset{1}

\if0\blind
{
\title{NACNet: A Histology Context-aware Transformer Graph Convolution Network for Predicting Treatment Response to Neoadjuvant Chemotherapy in Triple Negative Breast Cancer}
\author{Qiang Li$^{1}$, George Teodoro$^{2}$, Yi Jiang$^{1,3,*}$, Jun Kong $^{1,3,*}$  \\
$^{1}$Department of Mathematics and Statistics, Georgia State University, Atlanta, 30303, GA, USA\\
$^{2}$Department of Computer Science, Federal University of Minas Gerais, Belo Horizonte,\\ 31270, Minas Gerais, Brazil\\
$^{3}$Winship Cancer Institute, Emory University, Atlanta, 30322, GA, US}

}
\date{}
\maketitle
 \fi
		
\if1\blind
{
 \title{\bf \emph{IISE Transactions} \LaTeX \ Template}
\author{Author information is purposely removed for double-blind review}
			\bigskip
			\bigskip
			\bigskip
			\begin{center}
				{\LARGE\bf \emph{IISE Transactions} \LaTeX \ Template}
			\end{center}
			\medskip
} \fi

\spacingset{1.25} 

\noindent 
\begin{abstract}
\noindent Neoadjuvant chemotherapy (NAC) response prediction for triple negative breast cancer (TNBC) patients is a challenging task clinically as it requires understanding complex histology interactions within the tumor microenvironment (TME). Digital whole slide images (WSIs) capture detailed tissue information, but their giga-pixel size necessitates computational methods based on multiple instance learning, which typically analyze small, isolated image tiles without the spatial context of the TME. To address this limitation and incorporate TME spatial histology interactions in predicting NAC response for TNBC patients, we developed a histology context-aware transformer graph convolution network (NACNet). Our deep learning method identifies the histopathological labels on individual image tiles from WSIs, constructs a spatial TME graph, and represents each node with features derived from tissue texture and social network analysis. It predicts NAC response using a transformer graph convolution network model enhanced with graph isomorphism network layers. We evaluate our method with WSIs of a cohort of TNBC patient (N=105) and compared its performance with multiple state-of-the-art machine learning and deep learning models, including both graph and non-graph approaches. Our NACNet achieves 90.0\% accuracy, 96.0\% sensitivity, 88.0\% specificity, and an AUC of 0.82, through eight-fold cross-validation, outperforming baseline models. These comprehensive experimental results suggest that NACNet holds strong potential for stratifying TNBC patients by NAC response, thereby helping to prevent overtreatment, improve patient quality of life, reduce treatment cost, and enhance clinical outcomes, marking an important advancement toward personalized breast cancer treatment.
\end{abstract}
\renewcommand{\thefootnote}{\fnsymbol{footnote}}
\noindent \footnotetext[1]{Correspondence author: \\Email:yjiang12@gsu.edu (Y. Jiang); jkong@gsu.edu (J. Kong)}
\clearpage
\newpage

\bibliographystyle{unsrt}
\maketitle
\section{Introduction} \label{s:intro}
Breast Cancer is the leading malignant disease and cause of cancer death in women worldwide~\cite{boyle2012triple}. Of all breast cancer subtypes, Triple Negative Breast Cancer (TNBC) presents the most aggressive progression, with a dismal prognosis and a high recurrence rate~\cite{boyle2012triple}. In current clinical practice, treatment option for this cancer subtype is limited. Although new immunotherapies have recently emerged, chemotherapy remains the primary treatment for patients with TNBC at both early and advanced stages. In particular, because of the absence of estrogen receptors (ER), progesterone receptors (PR), and human epidermal growth factor receptor 2 (HER2), TNBCs do not respond to hormone therapies or HER2-targeted treatments, significantly narrowing the range of therapeutic options. 
Consequently, chemotherapy, particularly neoadjuvant chemotherapy (NAC),  continues to be the standard-of-care and predominant treatment option for this aggressive cancer subtype~\cite{hirano2023tissue,dai2023molecular,mehanna2019triple,valencia2022immunotherapy,bianchini2016triple}.

NAC treatment response is evaluated in resected tissues from the surgery using the residual cancer burden (RCB) metric. Based on RCB scores, two response classes are defined: Pathology Complete Response (pCR) with a post-treatment RCB of zero, and Residual Disease (RD), indicating an incomplete response. Clinically, pCR is often used as an endpoint for reoperative treatment and is strongly correlated with long-term clinical benefits~\cite{cortazar2014pathological,pennisi2016relevance}. 

However, only about 30\%$\sim$40\% of TNBC patients respond well to NAC treatment, while the remaining patients either respond moderately or are refractory to NAC~\cite{franceschini2018conservative}. Despite its clear clinical benefits, for example a reduction in the tumor size or a downgrade in breast cancer before the surgery, NAC can substantially decrease patient life quality with its adverse treatment effects~\cite{sakuma2011pathological}. For non-responders, NAC not only introduces unnecessary toxicity but also delays alternative treatments and surgery, leading to adverse outcomes and overtreatment. Unfortunately, there remains a critical, unmet need to accurately predict NAC treatment response at the time of diagnosis. An accurate treatment response prediction remains a major clinical challenge.

In recent years, machine learning has been successfully applied to histopathology images for predicting survival outcomes and treatment responses in the clinical oncology~\cite{wang2024pathology,ahn2024histopathologic}. Numerous studies have demonstrated the effectiveness of image-based machine learning in improving clinical decision-making. For example, deep learning models have been used to predict early treatment response in metastatic colorectal cancer by identifying subtle morphological changes in tumors, surpassing traditional size-based assessment methods~\cite{lu2021deep}. In breast cancer research, deep learning frameworks have been used to integrate histopathology images with additional data types, such as genomic and radiomic features, to predict NAC response~\cite{hoang2024deep,lee2024prediction,moon2023machine}. These models provide non-invasive methods for assessing treatment efficacy and guiding personalized therapy strategies. The successful applications of deep learning across various cancer types suggest its potential to revolutionize clinical treatment response prediction, offering more accurate and personalized approaches to support patient care.

Thanks to significant advances in tissue scanning technology, whole slide images (WSIs) of tissue sections can now be routinely produced to capture cell-level tissue details. However, due to their giga-pixel scale, WSIs pose significant computational challenges, making it difficult to directly deploy machine learning and image analysis techniques for large-scale automated analyses.~\cite{kumar2020whole,dimitriou2019deep,stritt2020orbit}. To manage the computational demand, it is a common practice to partition giga-pixel WSIs into numerous small image tiles and analyze them using a multiple instance learning (MIL) strategy~\cite{duanmu2020prediction,lee2020prediction,shin2012prediction,qu2020prediction,chen2021machine,zheng2022graph}. In MIL, tiles from each WSI are treated as instances from a bag that share the same patient-level label~\cite{andrews2002multiple,chen2021multimodal}. Each tile is assigned a predicted label and these tile-level labels are then aggregated to form a comprehensive patient-level representation~\cite{cruz2014automatic,lecun1998gradient,nguyen2009weakly}. However, this MIL strategy may be less effective due to the high tissue heterogeneity within the breast cancer tumor microenvironment (TME), which breaks the MIL assumption that all instances in the negative (RD) bag should have a negative label~\cite{bianchini2022treatment}. 
Notably, TNBCs exhibit significant TME heterogeneity, where the presence of tumor-infiltrating lymphocytes are associated with a better prognosis and an improved response to chemotherapy ~\cite{bianchini2016triple}. Numerous studies further highlight that TME histological interactions strongly correlated with NAC response in TNBC patients~\cite{bianchini2022treatment,keren2018structured,lycke2015mucosal,coluzzi2008overview}. However, MIL-based analyses ignore these spatial TME histology distributions and interactions, remaining insensitive to local tissue context and the global structure organization.

Graph Convolution Network (GCN) has been developed to partially address this problem. It represents each WSI as a graph consisting of nodes and edges, and predicts the WSI label by aggregated features of nodes in local neighborhoods~\cite{li2018graph,levy2020topological,chen2021whole,zheng2022graph}. Learning a structured graph requires an effective graph representation~\cite{hamilton2017}. However, constructing a context-aware, effective, and minimal WSI topological representation to hence predictive discriminating power remains an open challenge. Additionally, there is limited research on how to generate biologically meaningful graph node representations and edge connections for cancer grading and subtyping tasks using histopathology WSIs. Most current approaches cutting WSIs into tiles, then develop GCNs at the tile level, and finally aggregate the outcomes from the tile nodes to create overall WSI-level predictions~\cite{zheng2022graph,lu2022slidegraph+}. Although tile-level approaches have advanced research in WSI processing, they fall short in constructing WSI graphs with well-informed node labels and attributes. When the objective is to identify the entire tumor region or capture the TME connectivity in WSIs, where nodes characterize disease stage, it becomes important to incorporate both regional and WSI-level information with accurately labeled nodes and detailed attribute features. Therefore, label-informed WSI graph structure learning methods are needed for these prediction and analysis. 


To address the challenges posed by spatial heterogeneity in TNBC WSIs, we developed a multi-step NAC prediction network (NACNet) that leverages a WSI-derived graph representation and a transformer-based graph convolution network (GCN) for NAC response prediction in TNBC patients. Our approach leverages clinically relevant TME information such as cellular (e.g., lymphocyte infiltration) and extracellular matrix (e.g., collagen) heterogeneity for an accurate treatment response prediction.
First, we train a deep learning model to recognize local histology labels and produce histology label maps for WSIs. Next, we construct a spatial TME graph using tile-level histology label map, extract context aware graph features, and predict the NAC treatment response using a transformer GCN model that is boosted with Graph Isomorphism Network (GIN) layers~\cite{xu2018powerful}. We applied the NACNet to a TNBC patient cohort, and   demonstrated that incorporating accurate spatial TME structures enhanced NAC response prediction. 

\section{Dataset}
This study used a set of specimens from a TNBC patient cohort receiving NAC treatment at the Decatur Hospital, Georgia, USA. All patient tissues were histologically processed and stained with Hematoxylin and Eosin ($H\&E$). The H\&E-stained tissues were scanned using a high-throughput high-resolution digital scanner (Hamamatsu NanoZoomer 2.0-HT C9600-13) at 40x magnification (0.23 $\mu$m/pixel). Board-certified pathologists annotated histological features in these slides using an open-source image processing software (QuPath, ver. 0.1.2). The features include adipose tissue, Polyploid Giant Cancer Cells (PGCCs), Microvessel Density (MVD), tumor, stroma, Carcinoma In-Situ (CIS), hemorrhage, mucinous change, apocrine change, immune cells, necrosis, and muscle tissue. The TNBC patient cohort included 105 female patients in total, with 48 patients responding (pCR) and 57 not responding (RD) to the NAC treatment. 

\section{Methods} 
\begin{figure*} [ht]
	\centering
\includegraphics[width=\linewidth]{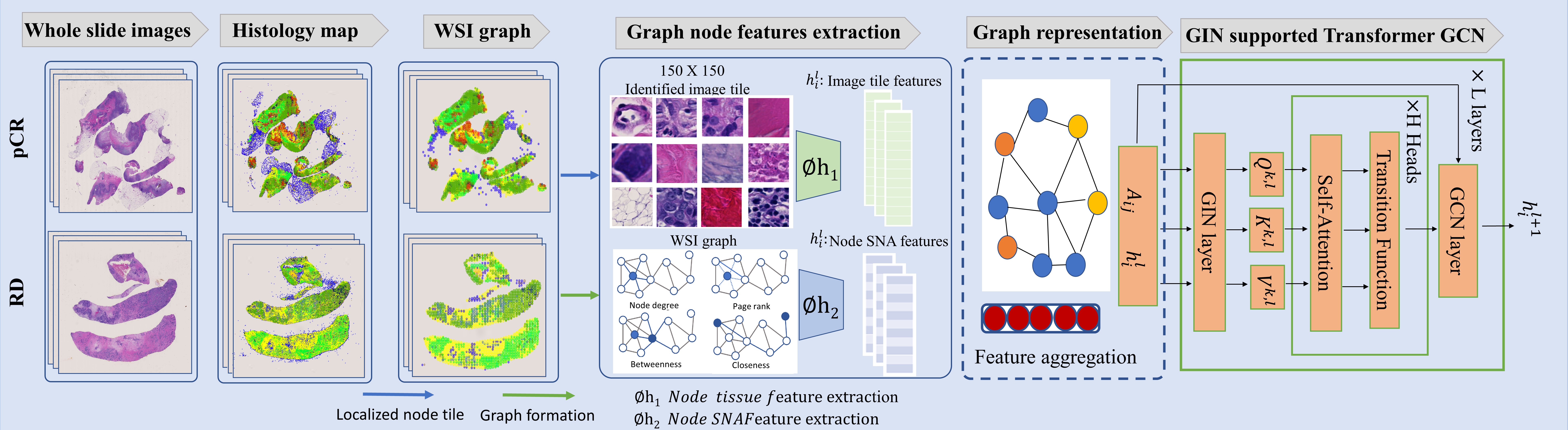}
   \caption{\textbf{Overview of the NACNet architecture.} Each WSI is partitioned into non-overlapping image tiles of size $150 \times 150$ pixels.A pre-trained convolutional model for image tile classification identifies all image tile classes, resulting in a tile-level histology label map where each pixel represents the histology label of an image tile. A sliding window moves over each histology label map and defines neighbors that share the same histology label as a WSI graph node. Both graph-based SNA and image texture features are used to represent the graph nodes. Nodes within the distance $\epsilon$ are connected with an edge. The resulting $\epsilon$-neighborhood graph is provided to a transformer-based GCN with GIN layers. This architecture incorporates  the TME spatial information, improving the prediction power.}~\label{fig:1}
\end{figure*}

We developed a histology context-aware transformer graph convolution network, NACNet, for an enhanced TNBC NAC response prediction with H\&E WSIs. NACNet recognizes histology labels of image tiles, constructs spatial TME graphs, characterizes graph nodes with the local node label information including both type and count, tissue texture, and social network analysis (SNA) features. We then used a transformer graph convolution network (GCN) equipped with graph isomorphism network (GIN) layers to analyze these graphs and predict the treatment outcome(Figure~\ref{fig:1}). 

\subsection{Histology map generation}
\begin{figure*}[!ht]
	\centering	\includegraphics[width=0.85\linewidth]{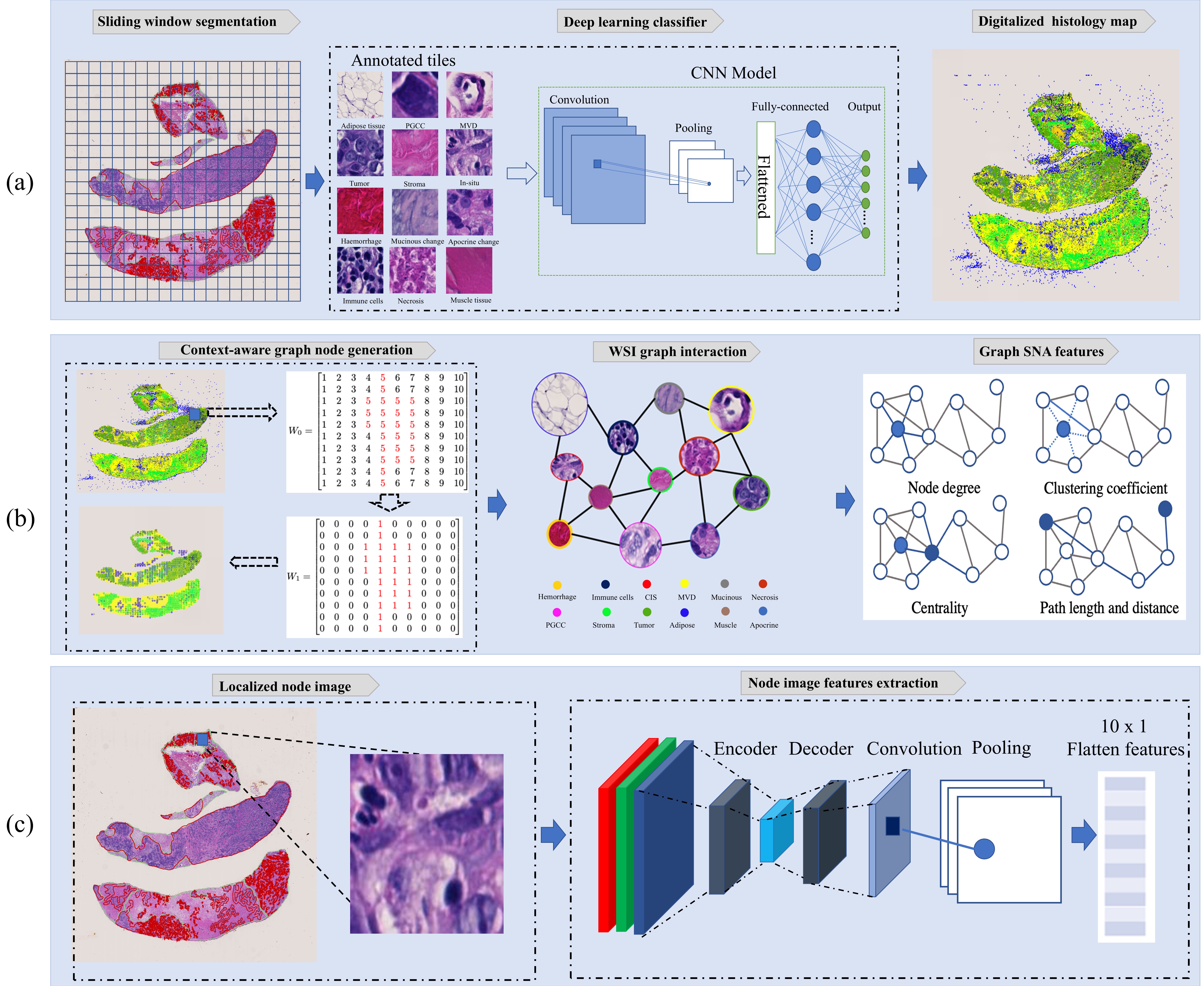}
\caption{\textbf{Feature extration from WSI spatial TME graph.} (a) Each WSI is partitioned into image tiles. A training set of slides are annotated with one of 12 histology classes. A VGG16 model~\cite{VGG16} trained on these annotated slides predict the histology labels for the rest of the titles. All labeled tiles are then combined to produce the histology map for each WSI. (b) Within the histology map, a large cluster of tiles ($n>\eta$) sharing the same histology label is defined as a TME graph node. Any pair of nodes within distance $\epsilon$ are connected by an edges. Multiple SNA features are derived from each graph node (e.g., degree, betweenness, page rank, and closeness). (c) An autoencoder connected to a VGG flatten layer is used to find image embeddings of graph nodes. The node size, tissue texture and SNA features are integrated to enable a context-aware graph characterization.}~\label{fig:2}
\end{figure*}

To accommodate computational and memory constraints, we partitioned each WSI into non-overlapping 150 $\times$ 150 image tiles and recorded the spatial coordinates of the top left corner of each tile. We trained a convolutional neural network (CNN) classifier to converts each tile into a histology label. Then a WSI became a histology label map (Figure~\ref{fig:2}a). 
The histology labels of training image tiles were based on the detailed histology annotations of 12 distinct histology classes by pathologists. Background regions are identified by the cutoff intensity value 220 in all RGB image channels. Tiles with more than 90\% background regions are discarded. Thanks to the large number of TNBC related histology components of interest in this study, the histology map can capture diverse and intricate histological characteristics important for TNBC treatment response prediction.

\subsection{TME graph construction}
Characterizing the spatial histopathological heterogeneity is critical for both understanding cancer progression and for cancer subtyping, biomarker discovery, and treatment response prediction~\cite{duanmu2020prediction,yang2019guided}. Our NACNet model creates a graph-based TME representation that takes advantage of the spatial interactions across different tissue histology regions in a WSI (Algorithm~\ref{alg:1}). The graph, $G = (V, E)$ is undirected, where $V$ is the set of nodes and $E$ is the set of edges connecting adjacent nodes.
\begin{figure*}[ht]
    \centering
    \begin{minipage}{\textwidth}
\begin{algorithm}[H]
\caption{WSI Graph Representation}\label{alg:1}
\begin{algorithmic}
\State \textbf{Input:} $W$: WSI; $M$: Histology map 
\State \textbf{Output:} $G$: Graph data; $L_{c}$: Subgraph clique distribution; $L_{e}$: Edge distribution 

\State \textbf{Stage 1:} Graph and Node Feature Generation
\For{each sliding window $m$ in the $M$}
        \State $C \gets$ SpatiallyClusterLabels($m$) \Comment{Spatially cluster histology map labels within the sliding window}
        \For{each cluster $c$ in $C$} 
         \State $label\_count \gets$ CountCluster($c$) \Comment{Count number of labels in the cluster}
     \If{$label\_count > \eta$} 
           \State $V \gets$ CalculateCentroid($c$) \Comment{Calculate centroid of the cluster}  
            \State $F \gets$ autoencoder($W(\text{centroid}(c))$) 
            \Comment{Extract image texture features}
        \EndIf
\EndFor
\EndFor
\For{each node $v_i$ in $V$}
            \If{$d(v_{i}, v_{j})<\epsilon$} 
                \State $E_{ij} \gets (v_{i}, v_{j})$ \Comment{Create the edge}
                        \EndIf
    \EndFor
\State \textbf{return} $G(V,E,F)$ \Comment{Graph and its representation}

\State \textbf{Stage 2:} Subgraph Cliques and Edge Distribution
\State $L_{c}(v_{i}, v_{j}, v_{k}) \gets$ allcycles($G$, $MCL$, $\lambda=3$) \Comment{Find all cycles of length three}
\State $L_{e}(v_{i}, v_{j}) \gets$ allcycles($G$, $MCL$, $\lambda=2$) \Comment{Find edges with all two-node histology label combinations}
\end{algorithmic}
\end{algorithm}
    \end{minipage}
\end{figure*}

As each image tile in our study is $150\times 150$ by size, we use a 10$\times$10 sliding window over the histology map to identify context-aware graph nodes by connecting neighboring tiles of the same histology label.  As we move the sliding window over the histology map, we spatially cluster the histology map labels within the sliding window. Each resulting cluster is defined as a graph node with its location represented by the cluster centroid.
We carefully considered both the physical size differences among histological categories and their clinical significance in breast cancer. Our approach to node formation is uniquely context-aware, aiming to capture the essential pathological features that inform prognosis and treatment strategies. Among our 12 histology categories, several—including tumor, necrosis, immune cells, microvessel density (MVD), stroma, and polyploid giant cancer cells (PGCC)—carry high clinical relevance due to their direct association with tumor progression, prognosis, and treatment response \cite{hanahan2011hallmarks, denkert2010tumor, folkman2002role}. For these critical categories, we assigned a lower threshold $n=5$ value, ensuring they cluster more readily into nodes and are prominently represented within the TME graph. This approach emphasizes the spatial distribution of these important features within the tissue structure, enhancing the model’s ability to focus on key regions that reflect the tumor microenvironment’s complexity. Conversely, categories with lower clinical impact, such as hemorrhage, mucinous changes, carcinoma in situ (CIS), adipose tissue, muscle tissue, and apocrine change, were assigned higher thresholds $n=10$. This prevents over-segmentation of less significant features while still allowing these elements to provide meaningful context within the overall histological landscape. This clinically-weighted clustering strategy not only reduces the impact of misclassifications but also enhances the model's accuracy in constructing a context-aware TME graph. By assigning weights based on clinical relevance, our approach uniquely captures the heterogeneous spatial and clinical characteristics of breast cancer tissues, thus offering a more nuanced and informative representation for prognostic modeling.
Additionally, the number of histology map entries sharing the same histology label in the sliding window is the local node label count (Figure~\ref{fig:2}b). A pair of nodes with spatial distance less than $\epsilon=1,500$ pixels (0.25 $\mu$m/pixel)~\cite{mackenzie2022neural}. The adjacency matrix $A=[a_{ij}]$, where $a_{ij} = 1$ if an edge exists between node $i$ and $j$ and $a_{ij} = 0$ otherwise, captures the TME edge set $E$ and models histology component spatial interactions in TME. 

For each graph node, we use the local node label, node label count, SNA features, and node tissue texture features to capture the spatial histology interactions in the TME graph. We calculate the local node label count by the number of image tiles associated with each node. In addition, we calculate the SNA features~\cite{otte2002social} that include node degree, betweenness, pagerank, and closeness (Figure~\ref{fig:2}b). Specifically, the node degree (or degree centrality) is the number of edges connected to the current node. The betweenness centrality of a node is the sum of the ratio of the number of the shortest paths from node S to T through current node to the number of the shortest paths from S to T over all pairs of S and T. PageRank is a link analysis algorithm originally developed to measuring the relative importance of web documents within the World Wide Web~\cite{rogers2002google}. It assigns a numerical weight to each node in our spatial TME graph that is calculated iteratively as the sum of the ratios of the PageRank of neighboring connected nodes to the number of outbound links from these neighboring connected nodes. Closeness centrality measures the ability of a node to pass information in a graph and is calculated as the reciprocal of the average shortest path distance from the current node to all other nodes in the graph. To include tissue details for the node characterization, we extract node image tile texture features using an autoencoder flatten layer~\cite{liou2014autoencoder} (Figure~\ref{fig:2}c), resulting in a feature vector of length $12$.  

For each node, the label, label count, texture features, and SNA features together form the node feature vector $h_v$ of length $d$. The spatial TME graph feature matrix $H\in \mathbb{R}^{n\times d}$ consists all the features of all node $n$. The adjacency matrix $A$ captures the edge connections, while the graph feature matrix $H$ represents the graph node attributes. The adjacency matrix $A$ and the feature matrix $H$ together represent the tissue spatial TME graph from each WSI, providing a reliable and standardized method for treatment response prediction.

\subsection{NAC treatment response prediction based on the TME graph}
Our NACNet aims to improve the accuracy and efficiency of NAC response predictions for TNBC patients using the WSIs. To achieve this goal, NACNet model combines the transformer Graph Convolution Network (GCN)~\cite{nguyen2022universal} with the Graph Isomorphism Network (GIN) layer~\cite{xu2018powerful} to predict the treatment response. The GCN learns the abstract graph features through message passing, where feature vectors in the hidden layers are iteratively updated by aggregation of feature vectors from neighboring nodes~\cite{kipf2016semi}. To improve the performance, we combine GCN with a self-attention mechanism introduced by transformer~\cite{nguyen2022universal}. 

In our study, we further enhance the model WSI-level representation and discrimination power by a GIN layer with an Multilayer Perceptron (MLP) before the transformer GCN based spatial TME graph prediction module. This integration empowers the model to capture intricate tissue patterns and histology component relationships from the spatial tissue graph. Specifically, the GIN layer is known for data distribution learning in a low-dimensional space. This good merit enables GINs to discriminate different class-conditional graph structures, and map graph structures of same class to similar embeddings with a strong generalizability~\cite{yanardag2015deep}. In our model architecture, the output of GIN with Multilayer Perceptron (MLP) is further provided to a transformer with a self-attention mechanism. The aggregated features are next provided to a transition function that incorporates residual connections~\cite{he2016deep} and a normalization layer (LNorm)~\cite{ba2016layer}. Finally, the resulting latent graph embeddings are processed by a GCN layer. In this way, the graph embeddings are iteratively updated by the GIN layer, transformer and the GCN:

\begin{equation}
h'^{(k)}_{v} = MLP^{(k)} \left( (1+\alpha^{(k)}) \cdot h_{v}^{(k)} +\sum_{u \in \mathcal{N}(v)} h_{u}^{(k)} \right)
\end{equation}
\begin{equation}
\tilde{H}^{'(k)} = Attention\left(H'^{(k)}Q^{(k)},H'^{(k)}K^{(k)},H'^{(k)}V^{(k)}\right)
\end{equation}
\begin{equation}
H^{(k+1)} = GCN(\mathbf{A},\tilde{H}^{'(k)})
\end{equation}
\noindent where $\alpha$ is a learnable parameter for the node representation update~\cite{xu2018powerful}; $h_{v}^{(k)}$ is the feature vector of node $v$ at the $k$-th layer, while $H^{(k)}$ is the graph feature matrix. For each node $v \in \mathcal{V}$, the feature vectors of a set of neighbors (i.e. $\mu \in \mathcal{N}(v)$) by the adjacency matrix $\mathbf{A}$ are uniformly sampled and aggregated with the current node feature vector. Additionally, $V^{(k)}, Q^{(k)}, K^{(k)} \in \mathbb{R}^{d\times d }$ are the value projection weight,  query-projection, and key-projection matrix in the transformer, respectively.

\subsection{TME histology component spatial interaction characterization}
As the complex spatial TME component interactions, e.g. the interactions across tumor, stromal and immune cells, can shed much insight on the underlying disease status, we include multiple TME histology components in the tissue WSIs within a spatial TME graph structure. With this method, we can readily extract TME features of treatment response predictive value and compare the spatial TME component interaction differences between NAC treatment responder (i.e., pCR) and non-responder (i.e., RD) group.

First, we analyze the group differences in the number of subgraph cliques associated with different histology class combinations. To make our analysis amenable, we count the 3-node subgraphs, as outlined in Algorithm~\ref{alg:1}. We initially identify all subgraphs of cycle length three in the spatial TME graph $G$ by the function \textit{allcycles}, with a Maximum Cycle Length (MCL) set to three. Next, we incrementally count the three-node subgraph clique associated with each of the $C^l_{3}$ histology class combinations of interest. Next, we identify and count edges connecting two nodes of all histology class label combinations of interest in the same way, but with a MCL set to two. Finally, a statistical t-test is used to assess the discrepancy in subgraph counts between the NAC responder and non-responder group.

In addition, we investigate the predictive value of the TME histology class profile for NAC response prediction.  Specifically, we calculate correlation coefficients between histology label count and NAC response (i.e. responders or non-responders) by Pearson correlation and produce the histology label correlation maps to intuitively demonstrate the importance of histology components. Additionally, we compute relevance scores by Minimum-Redundancy-Maximum-Relevance (mRMR) analysis~\cite{ding2005minimum} for histology label relevancy importance maps. The histology label correlation and relevancy importance maps by Pearson correlation and mRMR can visually provide insights into the contribution of individual histology labels to the prediction.

\subsection{Implementation and evaluation}
A VGG16 model for image tile classification: All image tiles are extracted from WSIs with the Python-Openslide package. For image tile classification, the VGG 16 model implemented with TensorFlow Keras 2.3.0 is trained and tested on a workstation equipped with an NVIDIA GTX 3080 GPU. We randomly select 80\% and 20\% of the labeled image tiles with similar prior histology label distributions for effective training and testing, respectively. For training, we employ a stochastic gradient descent optimizer with a learning rate of 0.001, Mini-batch size of 32, and a dropout rate of 0.2.

Autoencoder for image tile texture extraction: We implement an autoencoder architecture with a VGG16 convolutional layer for image tile texture extraction in our study. The autoencoder consists of an encoder and sequential decoder. The encoder compresses input data into a lower-dimensional latent space using densely connected layers with rectified linear unit activation. Conversely, the decoder reconstructs the original data from this compressed representation, employing a combination of dense layers and a reshape operation to match the input dimensions. The encoded representations are softmax-normalized. This approach enables both an efficient feature extraction and the feature dimensionality reduction.

Model for NAC response: We develop the spatial TME graph-based prediction model with the PyTorch framework. The resulting model is trained with the Adam optimizer~\cite{kingma2014adam}, a maximum of 250 epochs and a dropout rate of 0.5. Different learning rates are tested, including ${e^{-4}, 5e^{-4}, 1e^{-3}, 5e^{-3}}$. Two attention heads, two MLP layers, and a hidden size of 256 are included in the model (Supplement TableS1).

Evaluation metrics: We separate patients into two groups: those with pathological complete response (pCR) as the postive group and those with residual disease (RD) as the negative groups. Accuracy (ACC) measures the overall performance of the method, while precision reflects the model's ability to avoid false positives. The area under the ROC curve (AUC-ROC) evaluates the model's discrimination power across various classification thresholds. The F1 score provides a balanced measure, considering both precision and recall. We use an 8-fold cross-validation for the performance evaluation. In each fold, 87.5\% and 12.5\% of the patients are used for training and testing, respectively. The mean and standard deviation of testing accuracies from these 8 folds are calculated to assess the overall prediction performance. With the same training and validation schema, we evaluate numerous state-of-the-art methods for a fair comparison.

\section{Results}
\subsection{VGG16 model for histology map generation}
\begin{figure*}[!htbp] 
	\centering 
\includegraphics[width=0.65\linewidth]{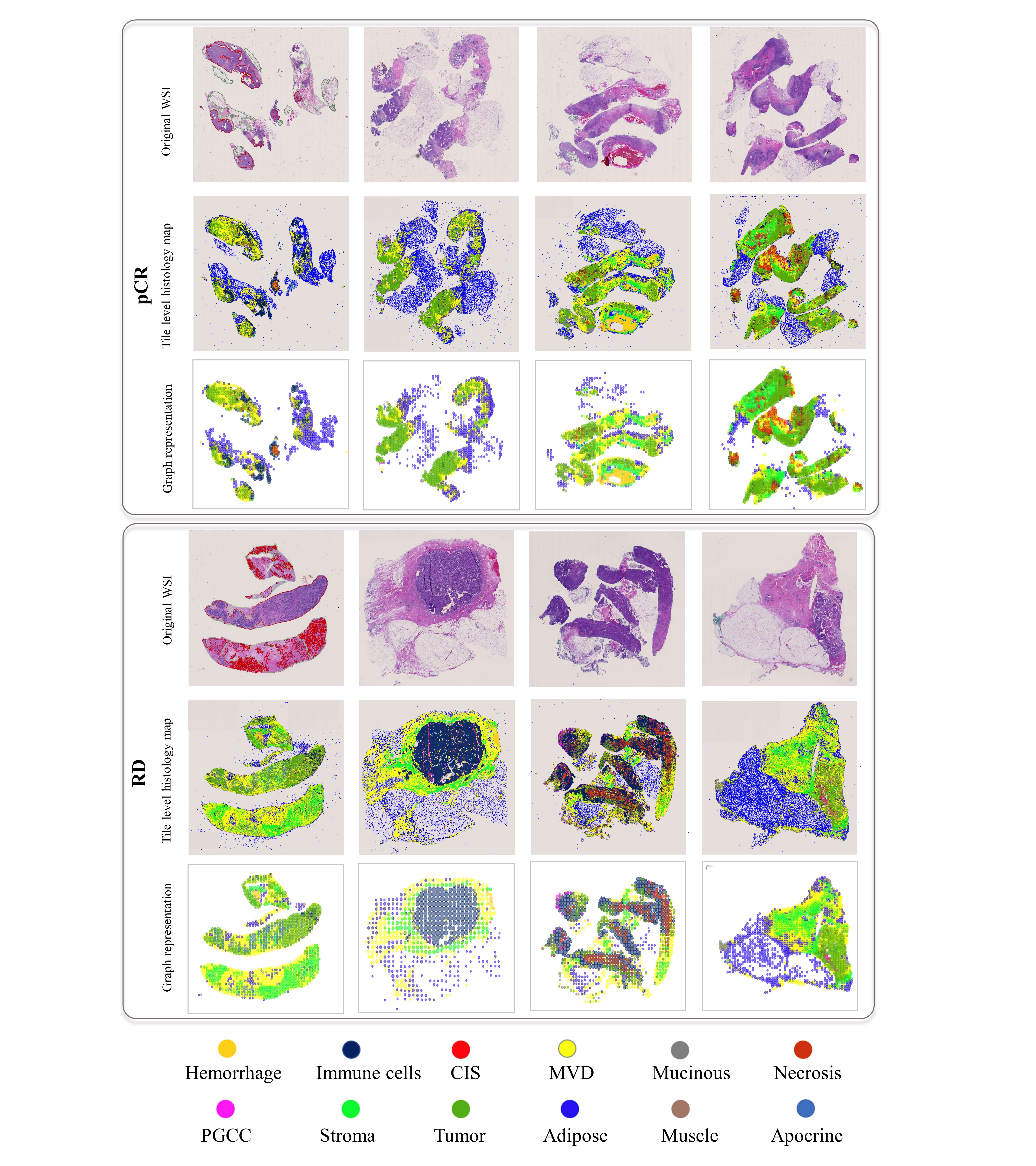}
\caption{\textbf{WSI-derived spatial TME graph.} We present representative WSIs (top), the corresponding histology maps (middle), and the resulting WSI TME graph node distributions (bottom). The histology labels of image tiles of size $150 \times 150$ are classified and assembled to construct the histology map. In total, 12 histology labels are color coded, including hemorrhage, immune cells, carcinoma in situ (CIS), MVD, mucinous changes, necrosis, PGCC, stroma, tumor, adipose tissue, muscle tissue, and apocrine change.}~\label{fig:3}
\end{figure*}

With a dataset of 10,000 annotated image tiles, the VGG16 model for image tile histology classification achieves a test accuracy of 90.0\%. We visually illustrate representative image tile histology maps and the WSI-derived spatial TME graphs (Figure~\ref{fig:3}). Through this implementation, we validate the performance of the VGG16 model in generating histology maps, laying a solid foundation and offering valuable insights for further tissue histology interaction analysis. By using the clustering method to identify nodes, we effectively prevent contamination of the resulting TME graph by image noise, such as tissue scratches, ink dots, dust marks, and fingerprints. 

\subsection{Ablation study and validation}
To evaluate the effectiveness of individual NACNet components, we first investigate the impact of individual WSI-derived spatial TME graph node features on the NAC response model performance using 8-fold cross-validation. Next, we examine NACNet's predictve performance by integrating multimodal node features. Both models with single- and multi-modal features are analyzed within the same NACNet pipeline framework. (1) NACNet-I uses only node image tile texture features; (2) NACNet-Lcontains the node label and count; (3) NACNet-S uses node SNA features; (4) NACNet-LI uses the node label, count, and node image tile texture features by autoencoder linked with a VGG flatten layer; (4) NACNet-LS uses node label, its count, and node SNA features; (5) NACNet-IS uses node image tile texture features and node SNA features; (6) NACNet-LIS includes node label, label count, node image tile texture features, and SNA features. 

\begin{figure*}[th]
	\centering
		\includegraphics[width=1\linewidth]{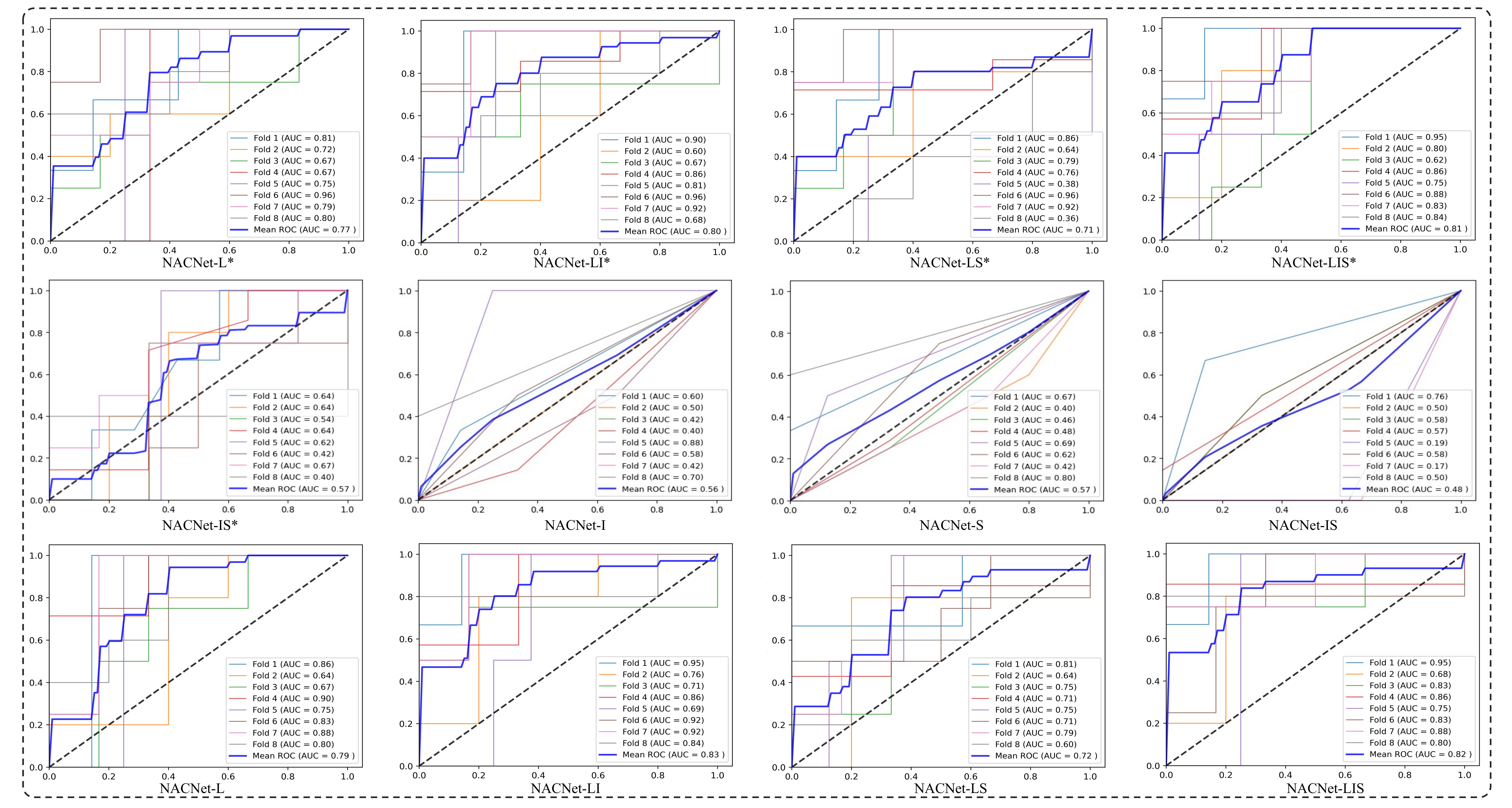}
\caption{\textbf{ROC curves of NACNet models in the ablation study.} The ROC curves for the NACNet methods are generated for each cross-validation fold. $\ast$ denotes models without the GIN layer in the network.}~\label{fig:4} 
\end{figure*}

We present the ablation experimental results with evaluation metrics of Accuracy, AUC, Sensitivity, Specificity, Precision, and F1 score in Table~\ref{tab:ablation}. The ROCs associated with different ablation models are illustrated in Figure~\ref{fig:4}. (1) Note the NACNet-LIS achieves the best prediction performance by Accuracy, and Sensitivity. For other metrics, it also presents competitive results. These results suggest the integrative use of multi-modal features help improve NAC response prediction. (2) Recall the image tile feature embeddings are produced by a convolutional auto-encoder and a VGG flatten layer. NACNet-LI works better than NACNet-LS overall, indicating the stronger complementary prediction power from image tile feature embeddings than SNA features. (3) Spatial node histology label and count information plays a critical role for NACNet model performance. Prediction models without node histology label and count information perform poorly. (4) Prediction models with the GIN layer work better than those without this layer overall, suggesting the efficacy of the GIN layer.

\begin{table*}[htbp]
\centering
\caption{Performance of NACNet using different node feature combinations (Mean $\pm$ Standard deviation). Models with ($\ast$) do not have a GIN layer. NACNet-I uses texture features by an autoencoder and a VGG flatten layer; NACNet-L uses the node label count; NACNet-LI uses both node label count and image tile texture features; NACNet-LS uses both node label count and SNA features; NACNet-LIS uses node label count, image texture features, and SNA features.}~\label{tab:ablation}
\scalebox{0.85}{\begin{tabularx}{1\linewidth}{Xcccccc}
\toprule
\toprule
\textbf{Models} & \textbf{ACC} & \textbf{AUC} & \textbf{Sensitivity} & \textbf{Specificity} & \textbf{Precision} & \textbf{F1} \\
\midrule
NACNet-L* & $0.78\pm0.12$ & $0.77\pm0.17$ & $0.57\pm0.22$ & $0.93\pm0.09$ & $0.90\pm0.13$ & $0.67\pm0.16$ \\
\addlinespace
NACNet-LI* & $0.80\pm0.19$ & $0.81\pm0.14$ & $0.83\pm0.15$ & $0.78\pm0.25$ & $0.76\pm0.21$ & $0.79\pm0.18$ \\
\addlinespace
NACNet-LS* & $0.74\pm0.11$ & $0.71\pm0.22$ & $0.54\pm0.24$ & $0.87\pm0.13$ & $0.71\pm0.32$ & $0.60\pm0.25$ \\
\addlinespace
NACNet-LIS* & $0.81\pm0.14$ & $0.81\pm0.14$ & $0.78\pm0.20$ & $0.83\pm0.15$ & $0.77\pm0.19$ & $0.78\pm0.17$ \\
\addlinespace
NACNet-IS* & $0.48\pm0.10$ & $0.57\pm0.10$ & $0.10\pm0.14$ & $0.11\pm0.15$ & $0.14\pm0.19$ & $0.79\pm0.16$ \\
\addlinespace
NACNet-I & $0.55\pm0.17$ & $0.56\pm0.16$ & $0.47\pm0.23$ & $0.52\pm0.19$ & $0.45\pm0.12$ & $0.65\pm0.22$ \\
\addlinespace
NACNet-S & $0.59\pm0.18$ & $0.57\pm0.14$ & $0.48\pm0.16$ & $0.60\pm0.25$ & $0.49\pm0.13$ & $0.66\pm0.28$ \\
\addlinespace
NACNet-IS & $0.49\pm0.18$ & $0.48\pm0.19$ & $0.33\pm0.23$ & $0.46\pm0.31$ & $0.35\pm0.23$ & $0.64\pm0.21$ \\
\addlinespace
NACNet-L & $0.81\pm0.11$ & $0.79\pm0.16$ & $0.76\pm0.31$ & $0.78\pm0.11$ & $0.75\pm0.17$ & $0.80\pm0.13$ \\
\addlinespace
NACNet-LI & $0.88\pm0.08$ & $0.83\pm0.12$ & $\mathbf{0.96\pm0.09}$ & $0.83\pm0.15$ & $0.82\pm0.15$ & $0.91\pm0.08$ \\
\addlinespace
NACNet-LS & $0.79\pm0.09$ & $0.71\pm0.20$ & $0.62\pm0.27$ & $0.90\pm0.08$ & $0.74\pm0.31$ & $0.66\pm0.27$ \\
\addlinespace \midrule
\textbf{NACNet-LIS} & $\mathbf{0.90\pm0.07}$ & $\mathbf{0.82\pm0.11}$ & $\mathbf{0.96\pm0.08}$ & $\mathbf{0.88\pm0.10}$ & $\mathbf{0.83\pm0.15}$ & $\mathbf{0.88\pm0.09}$ \\
\bottomrule
\bottomrule
\end{tabularx}}
\end{table*}

\subsection{Performance comparison with SOTA}
\begin{figure*} [!ht]
	\centering
		\includegraphics[width=0.87\linewidth]{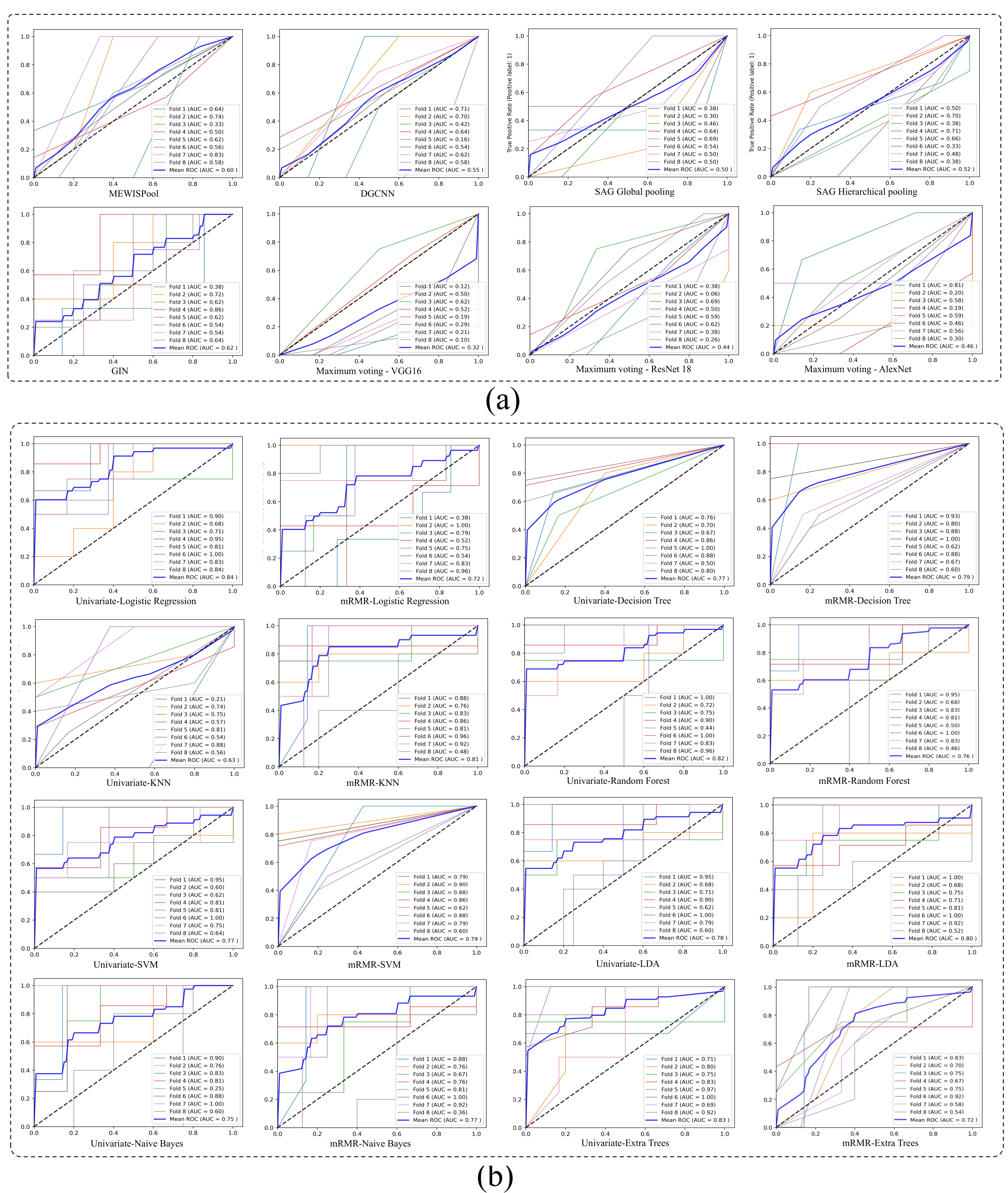}
\caption{\textbf{The ROC curves of methods for comparison.} The ROC curves of (a) 
deep learning methods and (b) traditional machine learning methods.}~\label{fig:5}
\end{figure*}

\noindent Comparison with GCN methods: 
Our NACNet model is compared with multiple state-of-the-art (SOTA) graph baseline deep learning methods, including Maximum Entropy Weighted Independent Set Pooling for GNNs (MEWISPool)~\cite{nouranizadeh2021maximum}, Deep Graph CNN (DGCNN)~\cite{zhang2018end}, SAGPool~\cite{lee2019self}, and SAG Hierarchical pooling~\cite{zhang2019hierarchical}. For fair comparisons, we use the same training hyperparameters, loss function, and eight-fold training and testing cross-validation partitions. We present the comparison results in Table~\ref{tab:sota} (top) and Figure~\ref{fig:5}(a). Our NACNet model with node information, image texture features and SNA features performs substantially better than the SOTA graph baseline GCN models by all evaluation metrics. Interestingly, most ablated NACNet models in Table~\ref{tab:ablation} outperform the SOTA graph baseline deep learning methods.

\begin{table*}[htbp]
\centering
\caption{Performance of deep learning models for NAC response prediction (Mean $\pm$ Standard deviation).}
~\label{tab:sota}
\scalebox{0.85}{\begin{tabularx}{1.1\linewidth}{Xcccccc}
\toprule
\toprule
\textbf{Models} & \textbf{ACC} & \textbf{AUC} & \textbf{Sensitivity} & \textbf{Specificity} & \textbf{Precision} & \textbf{F1} \\
\midrule
MEWISPool & $0.61\pm0.15$ & $0.68\pm0.19$ & $0.62\pm0.17$ & $0.68\pm0.12$ & $0.70\pm0.17$ & $0.64\pm0.12$ \\
\addlinespace
DGCNN & $0.64\pm0.14$ & $0.78\pm0.14$ & $0.62\pm0.16$ & $0.65\pm0.12$ & $0.74\pm0.14$ & $0.66\pm0.12$ \\
\addlinespace
SAG Global pooling & $0.64\pm0.11$ & $0.77\pm0.14$ & $0.59\pm0.10$ & $0.68\pm0.08$ & $0.71\pm0.09$ & $0.65\pm0.08$ \\
\addlinespace
SAG Hierarchical pooling & $0.63\pm0.11$ & $0.78\pm0.14$ & $0.64\pm0.17$ & $0.61\pm0.10$ & $0.69\pm0.15$ & $0.64\pm0.10$ \\
\addlinespace
GIN & $0.64\pm0.17$ & $0.74\pm0.15$ & $0.66\pm0.20$ & $0.63\pm0.15$ & $0.71\pm0.17$ & $0.67\pm0.15$ \\
\addlinespace \midrule
Maximum voting-VGG16 & $0.48\pm0.07$ & $0.42\pm0.20$ & $0.52\pm0.18$ & $0.36\pm0.15$ & $0.64\pm0.16$ & $0.54\pm0.15$ \\
\addlinespace
Maximum voting-ResNet 18 & $0.52\pm0.17$ & $0.53\pm0.18$ & $0.60\pm0.26$ & $0.42\pm0.20$ & $0.60\pm0.15$ & $0.58\pm0.20$ \\
\addlinespace
Maximum voting - AlexNet & $0.54\pm0.11$ & $0.52\pm0.15$ & $0.62\pm0.20$ & $0.39\pm0.13$ & $0.59\pm0.12$ & $0.59\pm0.13$ \\
\addlinespace ~\label{tab:sota}
\textbf{NACNet-LIS} & $\mathbf{0.90\pm0.07}$ & $\mathbf{0.82\pm0.11}$ & $\mathbf{0.96\pm0.08}$ & $\mathbf{0.88\pm0.10}$ & $\mathbf{0.83\pm0.15}$ & $\mathbf{0.88\pm0.09}$ \\
\bottomrule
\bottomrule
\end{tabularx}}
\end{table*}

Comparison with non-graph deep learning models: 
We use the multiple instance learning strategy for non-graph deep learning model training. Each image tile is labeled with the same NAC response for the patient. 
For testing, a majority voting mechanism is used to aggregate tile-level predictions into a final NAC response prediction at the patient level. Specifically, VGG-16, AlexNet, and ResNet~\cite{VGG16,krizhevsky2012imagenet,he2016deep} are trained and evaluated. The comparisons are presented in Table~\ref{tab:sota} (bottom) and Figure~\ref{fig:5}(a). While these methods present performance comparable to graph-based deep learning methods, they are substantially inferior to our NACNet models.

Comparison with traditional machine learning methods: 
We evaluate multiple traditional machine learning models using histology label proportion, and WSI-graph SNA features with the same 8-fold cross-validation mechanism. The traditional machine learning classifiersinclude Logistic Regression~\cite{LR}, k-Nearest Neighbors (kNN)~\cite{KNN}, Support Vector Machines (SVM)~\cite{SVM}, Naive Bayes \cite{NB}, Decision Trees~\cite{DT}, Random Forests~\cite{RD}, Extra Trees~\cite{ET}, and Linear Discriminant Analysis (LDA)~\cite{LDA}. A majority voting mechanism is used to aggregate tile-level predictions into a final NAC response prediction at the patient level. Specifically, these models are trained independently using univariate feature selection~\cite{pedregosa2011scikit} and mRMR feature ranking method. The univariate feature selection selects the best features based on univariate ANOVA F-test and is used to identify significant difference. The mRMR feature selection method aims to select features that are highly relevant to the target variable while minimizing redundancy among them. By balancing the trade-off between feature relevance and redundancy, this approach improves the model's predictive performance by focusing on the most informative and distinct features. Effective feature selection reduces data redundancy, minimizes dependency on noisy features in predicting NAC treatment response, mitigates overfitting, ultimately resulting in an improved model accuracy and robustness. The predictions are presented in Table~\ref{tab:ml} and Figure~\ref{fig:5}(b), showing that traditional machine learning methods work better than non-graph deep learning methods with the maximum voting strategy.  Although feature selection methods improve traditional machine learning model performance, the NACNet model presents a superior performance to traditional machine learning methods overall. This improvement suggests that incorporating node image tile texture features and local histology labels significantly boosts the model's discrimination power for NAC response prediction. Additional results supporting the statistical comparison of treatment prediction accuracy are presented in Supplementary Table S2.

\begin{table*}[th]
\centering
\caption{NAC treatment response prediction of machine learning models performance (Mean $\pm$ Standard deviation).}
~\label{tab:ml}
\scalebox{0.85}{\begin{tabularx}{1.1\linewidth}{Xcccccc}
\toprule
\toprule
\textbf{Models} & \textbf{ACC} & \textbf{AUC} & \textbf{Sensitivity} & \textbf{Specificity} & \textbf{Precision} & \textbf{F1} \\
\midrule
Univariate-LR & $0.75\pm0.11$ & $0.84\pm0.10$ & $0.48\pm0.23$ & $0.96\pm0.07$ & $0.93\pm0.12$ & $0.59\pm0.17$ \\
\addlinespace
mRMR-LR & $0.73\pm0.15$ & $0.72\pm0.21$ & $0.70\pm0.32$ & $0.70\pm0.15$ & $0.59\pm0.27$ & $0.63\pm0.29$ \\
\addlinespace
Univariate-KNN & $0.71\pm0.11$ & $0.63\pm0.20$ & $0.29\pm0.24$ & $\mathbf{1.00\pm0.00}$ & $0.63\pm0.48$ & $0.39\pm0.31$ \\
\addlinespace
mRMR-KNN & $0.71\pm0.12$ & $0.81\pm0.14$ & $0.68\pm0.18$ & $0.75\pm0.16$ & $0.65\pm0.19$ & $0.65\pm0.15$ \\
\addlinespace
Univariate-SVM & $0.73\pm0.16$ & $0.77\pm0.14$ & $0.48\pm0.25$ & $0.94\pm0.08$ & $0.85\pm0.19$ & $0.57\pm0.21$ \\
\addlinespace
mRMR-SVM & $0.79\pm0.11$ & $0.79\pm0.11$ & $0.71\pm0.17$ & $0.87\pm0.15$ & $0.78\pm0.25$ & $0.72\pm0.17$ \\
\addlinespace
Univariate-Naive Bayes & $0.76\pm0.13$ & $0.75\pm0.22$ & $0.81\pm0.32$ & $0.68\pm0.17$ & $0.63\pm0.26$ & $0.70\pm0.28$ \\
\addlinespace
mRMR-Naive Bayes & $0.73\pm0.15$ & $0.77\pm0.18$ & $0.80\pm0.18$ & $0.66\pm0.18$ & $0.64\pm0.17$ & $0.70\pm0.17$ \\
\addlinespace
Univariate-Decision Tree & $0.78\pm0.14$ & $0.77\pm0.14$ & $0.69\pm0.16$ & $0.85\pm0.19$ & $0.80\pm0.22$ & $0.73\pm0.16$ \\
\addlinespace
mRMR-Decision Tree & $0.81\pm0.13$ & $0.80\pm0.14$ & $0.69\pm0.21$ & $0.91\pm0.10$ & $\mathbf{0.90\pm0.13}$ & $0.72\pm0.20$ \\
\addlinespace
Univariate-Random Forest & $0.79\pm0.14$ & $0.83\pm0.18$ & $0.61\pm0.32$ & $0.92\pm0.17$ & $0.78\pm0.34$ & $0.65\pm0.30$ \\
\addlinespace
mRMR-Random Forest & $0.74\pm0.13$ & $0.76\pm0.18$ & $0.52\pm0.25$ & $0.90\pm0.13$ & $0.74\pm0.33$ & $0.59\pm0.26$ \\
\addlinespace
Univariate-LDA & $0.74\pm0.14$ & $0.77\pm0.14$ & $0.61\pm0.28$ & $0.82\pm0.18$ & $0.68\pm0.33$ & $0.63\pm0.28$ \\
\addlinespace
mRMR-LDA & $0.80\pm0.10$ & $0.80\pm0.16$ & $0.72\pm0.21$ & $0.88\pm0.11$ & $0.80\pm0.18$ & $0.73\pm0.15$ \\
\addlinespace
Univariate-Extra Trees & $0.61\pm0.15$ & $0.83\pm0.11$ & $0.47\pm0.25$ & $0.95\pm0.08$ & $0.79\pm0.34$ & $0.58\pm0.27$ \\
\addlinespace
mRMR-Extra Trees & $0.76\pm0.21$ & $0.72\pm0.12$ & $0.61\pm0.15$ & $0.84\pm0.12$ & $0.61\pm0.15$ & $0.61\pm0.15$ \\
\addlinespace
ReliefF-1NN & $0.73\pm0.07$ & $0.74\pm0.10$ & $0.95\pm0.09$ & $0.53\pm0.18$ & $0.60\pm0.09$ & $0.73\pm0.07$ \\
\addlinespace
ReliefF ensembleTree & $0.70\pm0.11$ & $0.69\pm0.10$ & $0.70\pm0.20$ & $0.53\pm0.18$ & $0.58\pm0.12$ & $0.68\pm0.14$ \\
\addlinespace
ReliefF rbf-SVM & $0.75\pm0.11$ & $0.62\pm0.10$ & $0.48\pm0.23$ & $0.47\pm0.23$ & $0.53\pm0.11$ & $0.62\pm0.12$ \\
\addlinespace
ReliefF lin-SVM & $0.73\pm0.15$ & $\mathbf{0.86\pm0.07}$ & $0.70\pm0.32$ & $0.74\pm0.13$ & $0.71\pm0.16$ & $0.82\pm0.11$ \\
\addlinespace \midrule
\textbf{NACNet-LIS} & $\mathbf{0.90\pm0.07}$ & $0.82\pm0.11$ & $\mathbf{0.96\pm0.08}$ & $0.88\pm0.10$ & $0.83\pm0.15$ & $\mathbf{0.88\pm0.09}$ \\
\bottomrule
\bottomrule
\end{tabularx}}
\end{table*}

\subsection{Spatial TME analysis by graph structures}
To enable a comprehensive understanding of the underlying biological mechanisms influencing NAC response, we compare the TME histology profiles between NAC treatment responder and non-responder group. The resulting group differences can manifest the impact of specific histology components on the NAC treatment outcome. Additionally, the distributions of the graph edges associated with different pairs of node histology labels are studied between these two patient populations.

\begin{figure*}[!ht]
	\centering
		\includegraphics[width=0.65\linewidth]{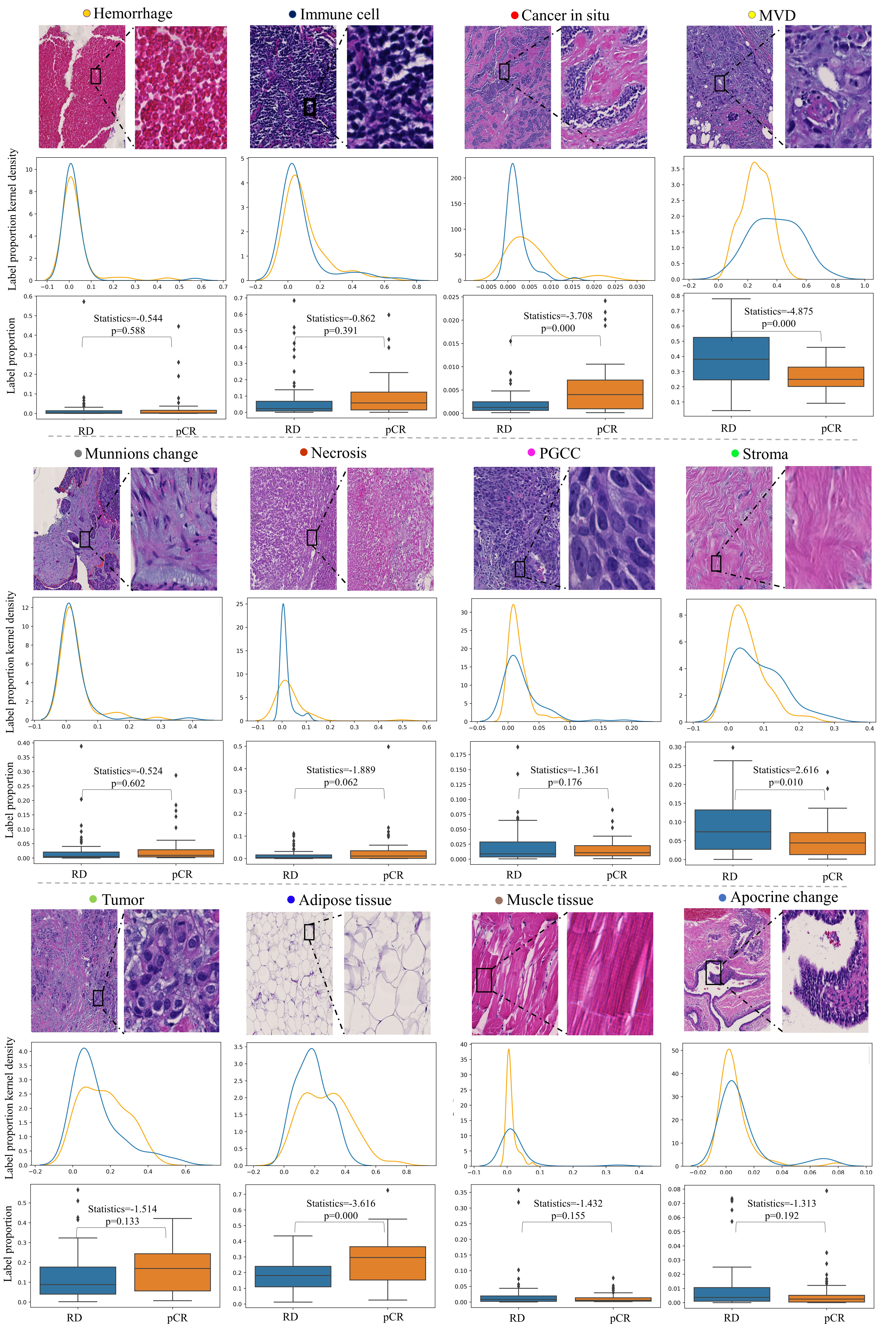}
\caption{\textbf{Comparison of histology profiles between pCR and RD patients.} We plot the individual histology label proportion distributions for both pCR (orange) and RD (blue) TNBC patients. A two-sample t-test is used to assess if individual histology label proportion means are significantly different between these two patient groups. Box plots present quartile values and whiskers extended to data points within the 1.53 interquartile range.}~\label{fig:6}
\end{figure*}

TME histology spatial features are well-established as critical factors influencing tumor growth, progression, and metastasis, as cancer cells interact with surrounding stroma and inflammatory cells~\cite{greten2019inflammation,jiao2021deep,amgad2023population}. To investigate their impact on TNBCs, we compare histology profiles between pCR and RD patients, and identify specific histology patterns potentially associations with NAC response. As shown in Figure~\ref{fig:6}, the proportions of CIS (p=0.000), MVD (p=0.000), stroma (p=0.010), and adipose tissue (p=0.000) differ significantly between the two patient groups. Notably, all WSIs in this study have a similar effective tissue scale, allowing for direct comparison of WSI graph edges and subgraph structures of interest across patient groups. 

\begin{figure*}[!ht]
	\centering
		\includegraphics[width=0.70\linewidth]{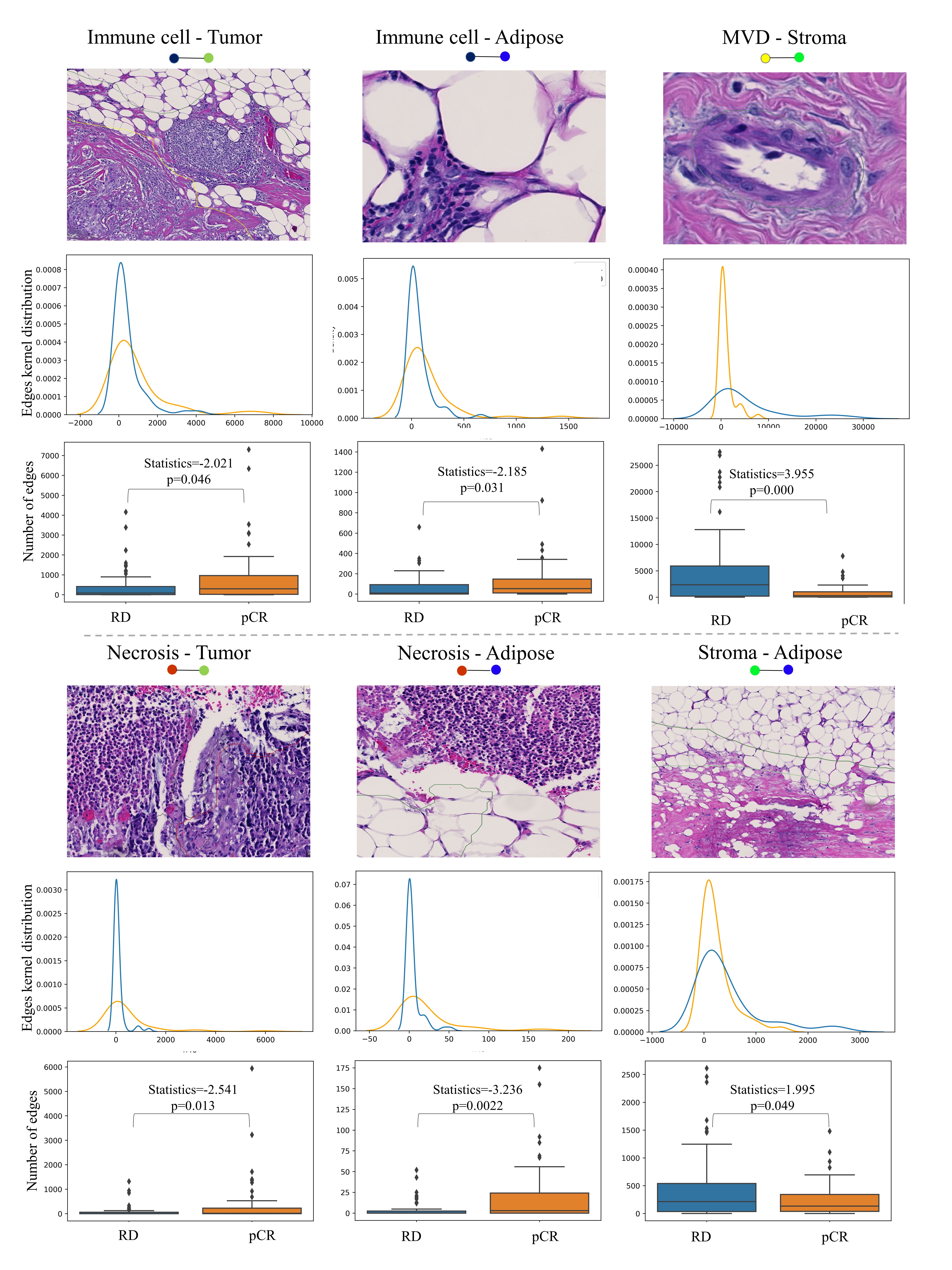}
\caption{\textbf{Comparison of WSI-derived graph edge profiles between pCR and RD patients.} As we include 12 histology labels in this study, there are 66 edge types associated with different histology label pairs. For each type, the edge counts were compared between pCR and RD patients. For each significant edge type, a typical tissue region associated with the significant edge type is illustrated. Additionally, we plot the individual graph edge distributions for both pCR (orange) and RD (blue) TNBC patients. A two-sample t-test is used to assess if individual graph edge means are significantly different between these two patient groups. Box plots present quartile values and whiskers extended to data points within the 1.53 interquartile range.}~\label{fig:7}
\end{figure*}

Figure~\ref{fig:7} shows histology pairs with significantly different mean graph edge counts between the two patient groups, based on t-test results. Edges associated with immune-tumor cells (p=0.046), immune-adipose (p=0.031), necrosis-tumor (p=0.013), and necrosis-adipose (p=0.0022) interactions are significantly more enriched in pCR group than in the RD group. Conversely, MVD-stroma (p=0.000) and stroma-adipose (p=0.049) edges are significantly more prevalent in RD patients. 

\begin{figure*}[!ht]
	\centering
		\includegraphics[width=1\linewidth]{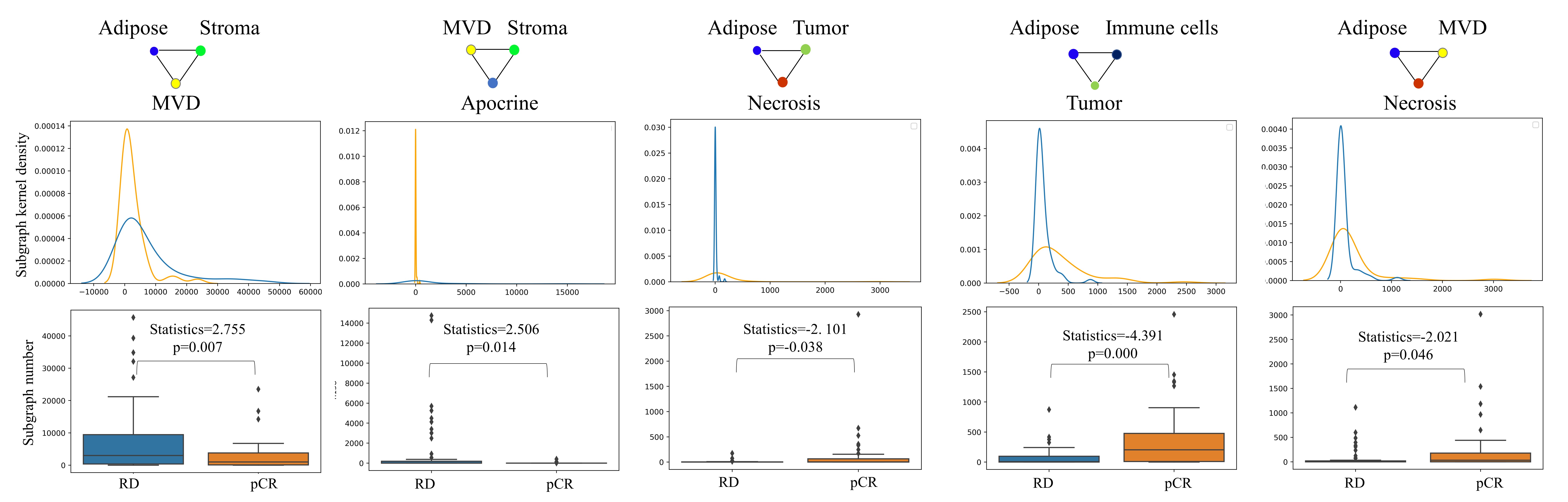}
\caption{\textbf{Comparison of WSI-derived subgraph profiles between pCR and RD patients.} 
As we include 12 histology labels in this study, there are 220 3-node subgraph types associated with different histology label triplets. We plot the individual 3-node subgraph distributions for both pCR (orange) and RD (blue) TNBC patients. A two-sample t-test is used to assess if individual 3-node subgraph means are significantly different between these two patient groups. Box plots present quartile values and whiskers extended to data points within the 1.53 interquartile range.}~\label{fig:8}
\end{figure*}

Next, we investigate the impact of 3-node subgraph quantification on NAC response, as shown in Figure~\ref{fig:8}. Subgraphs involving adipose-stroma-MVD (p=0.007) and MVD-stroma-adipose (p=0.014) are more enriched in RD patients. In contrast, subgraphs containing adipose-tumor-necrosis (p=0.038), adipose-immune cells-tumor (p=0.000), and adipose-MVD-necrosis (p=0.046) are significantly more enriched in the pCR group.

\begin{figure*} [!ht]
	\centering	\includegraphics[width=\linewidth]{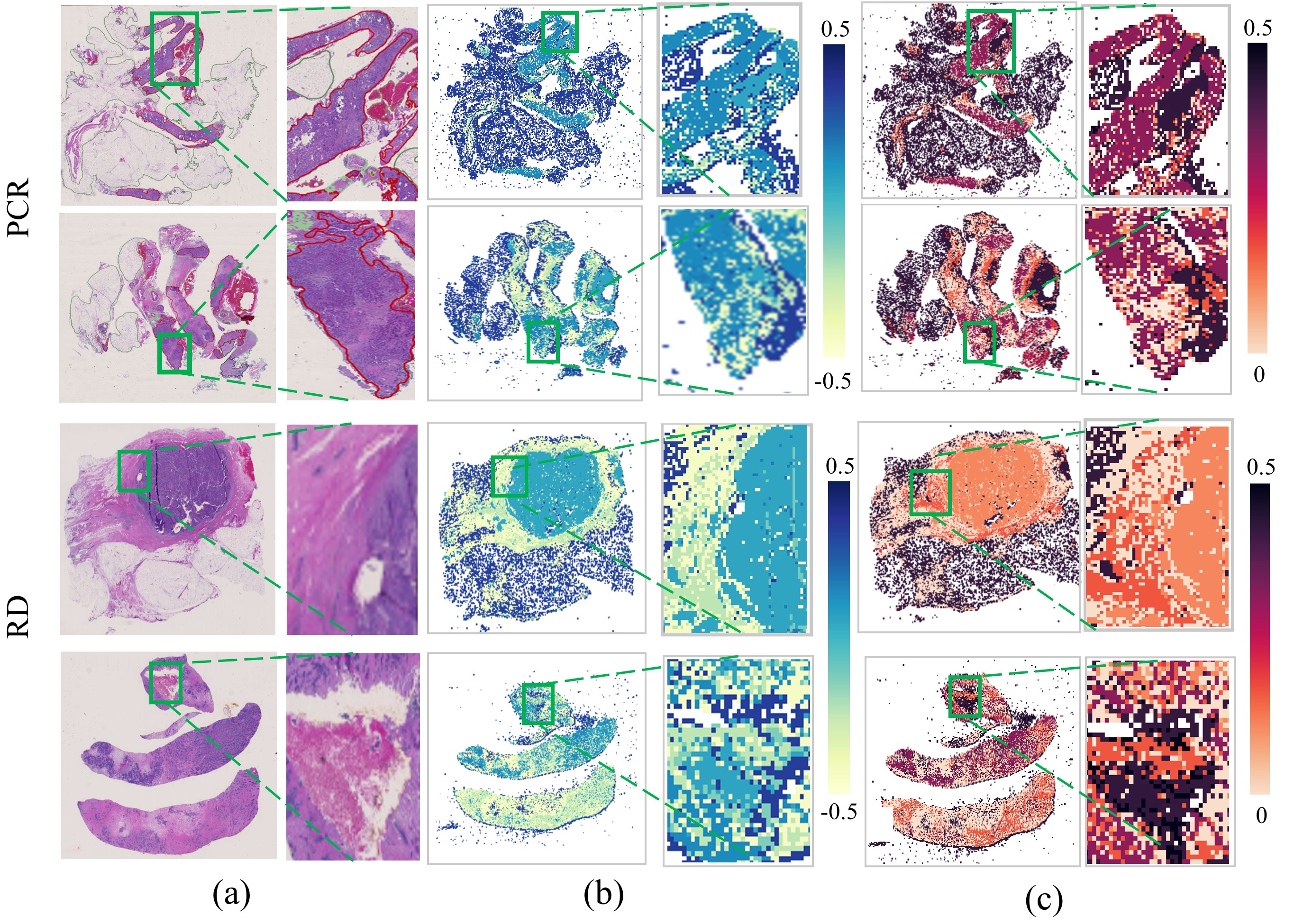}
\caption{\textbf{Prediction model interpretations with correlation and relevancy importance maps.} (a) Typical WSIs with their local tissue regions from pCR and RD group; (b) The histology label correlation map is produced by mapping to the histology map correlation coefficients by the Pearson correlation. The correlation value ranges from -1 to 1 and indicates the strength and direction of the relationship between individual histology labels and prediction outcomes; (c) The importance of graph feature nodes is analyzed using the mRMR (minimum Redundancy Maximum Relevance) method. The histology node importance map is generated by mapping to the histology map the importance values by the mRMR method.}~\label{fig:9}
\end{figure*}

We also generate histology label correlation maps, illustrating the Pearson correlation coefficients between histology label count and NAC response outcome, as well as histology label relevancy importance maps derived from mRMR analysis. These maps offer valuable insights into the decision-making process of the prediction model (Figure~\ref{fig:9}). 

\begin{figure*}[!ht]
	\centering
		\includegraphics[width=0.8\linewidth]{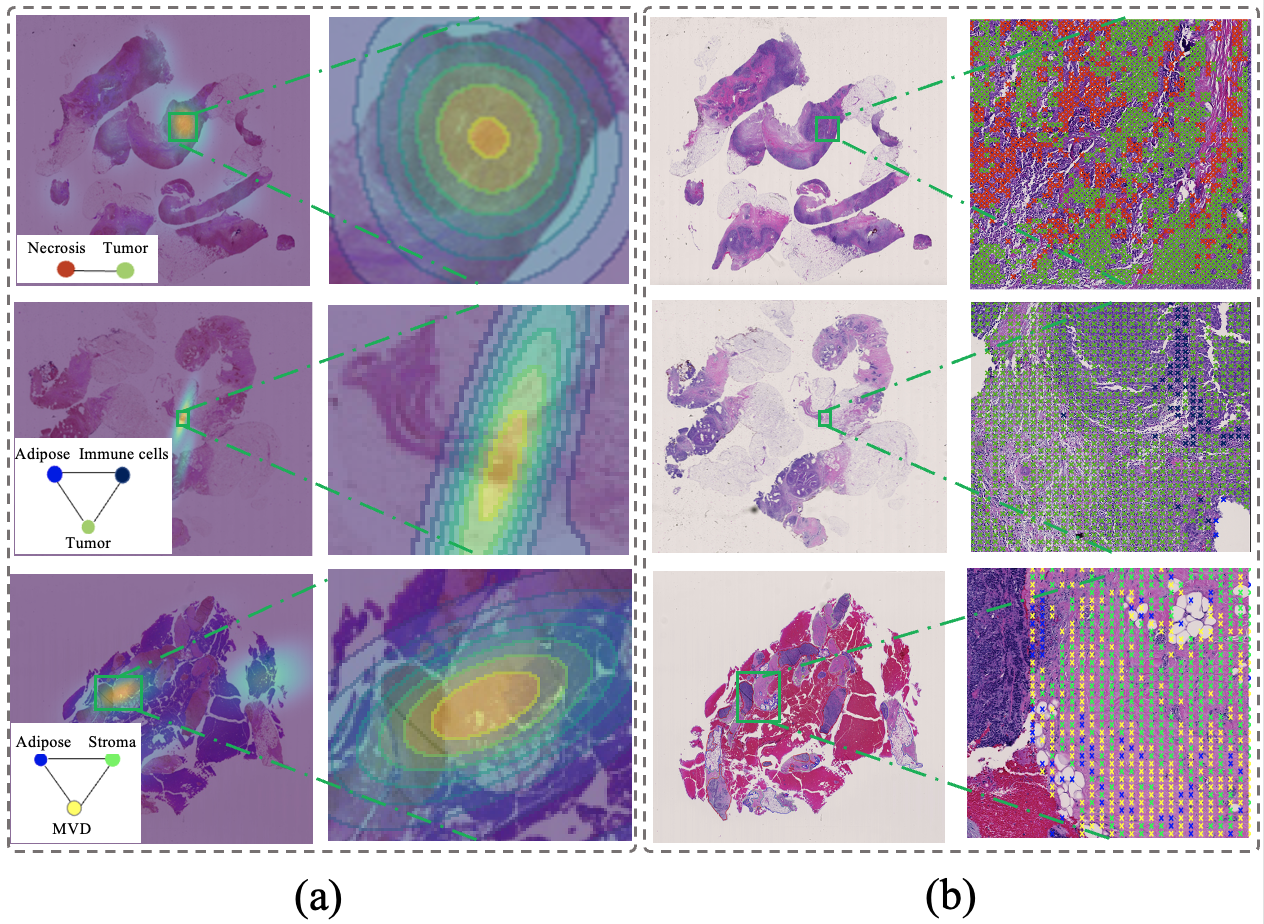}
\caption{\textbf{Attention visualization for typical tissue regions.} (a) We visualize spatial attention that targets specific subgraphs supporting the model prediction. The subgraph density is represented by iso-contours; (b) The histology labels of corresponding regions of interest are color coded by the same color schema in Figure~\ref{fig:3}.}~\label{fig:10}
\end{figure*}

Furthermore, we visualize in Figure~\ref{fig:10}(a) spatial attention that targets specific subgraphs used by the prediction model. The subgraph density is represented with iso-contours. In Figure~\ref{fig:10}(b), the corresponding tissue regions are overlaid with color coded histology labels by our model. Our pipeline unveils the intricate role of the TNBC TME by exploring spatial TME patterns, extracting local tissue texture information, and identifying potential biomarkers for NAC response prediction. Therefore, it is promising to enable an improved prognosis and treatment planning in practice, enhancing TNBC patient therapy and life quality. While our approach is developed for NAC treatment response prediction in TNBCs, it can be readily extended to other cancer types, contributing to a more comprehensive TME understanding.

\section{Discussion}
Deep learning has greatly advanced  computational pathology over the past decades. However, due to the vast scale of WSIs, much of the work has focused on decision support by processing small WSI tiles independently, overlooking crucial spatial connectivity within tissue. Our approach addresses this limiation by integrating global tissue graph structure signatures with local tissue texture and histology label information to enhance TNBC response prediction. 

Our contribution in this study can be summarized as follows. 

\begin{enumerate}
 \item We characterize histology features and their spatial interactions using TME graphs within a multi-step framework. First, we assign the histology label to each image tile using a deep learning module trained on $l=12$ histology classes. Next, we spatially assemble these labeled tiles based on their spatial arrangement, generating a tile-level histology label map for each WSI. Finally, we construct a spatial TME graph to capture the spatial histology organization critical for NAC response prediction. In this graph, each node represents the tile closest to the center of a cluster of tiles sharing the same histology label, while edges connect adjacent nodes based on spatial proximity.

\item To capture predictive features from both local tissue context and global tissue structure organization, we construct two complementary sets of features for graph node representation. First, an autoencoder is used to extract image texture features from tiles associated with each node. Second, social network analysis (SNA) features are computed to enhance node representation. This two-step approach integrates detailed local tissue information with spatial TME arrangement, providing a comprehensive foundation for accurate NAC response prediction.  

 \item NACNet is a transformer based GCN model designed to enhance the discriminating power of the graph-based prediction. To achieve this, we incorporate GIN layers into the graph transformer network. This layers transform heterogeneous input graphs into useful meta-path graphs and enable end-to-end learning of node representations. By  including the GIN layers, NACNet effectively captures the structure of WSI-derived TME graph, leading to improved prediction performance. 
\end{enumerate}

NACNet integrates histology label information, local tissue textures, and graph-derived SNA features to differentiate pCR from RD TNBC patients. By multiple evaluation metrics, NACNet achieves a better overall performance than state-of-the-art graph based deep learning architectures, non-graph deep learning models, and traditional machine learning approaches. NACNet efficiently captures local and WSI-level information to predict NAC response for TNBC patients. It processes high-resolution WSIs, predict multiple-class histology labels for local tissue regions, characterize TME spatial histology interactions, predict NAC response outcome, and generate interpretable findings associated with prediction.

Additionally, our work provides a straightforward way to enable the graph-based TME analysis. It can highlight the spatial histology patterns and the WSI regions that are significantly associated with the NAC response outcome. Unlike other graph-based classification models, NACNet is able to produce the TME attention over WSIs. This is useful as subgraphs and edges connecting different histology regions of interest in the neighborhood can contain biologically interpretable information that is critical for prediction. We perform detailed subgraph analysis and identify edges and three-node cliques of interpretation and prediction value. Such analyses enable biological interpretations especially important in histopathology WSI analysis applications where the detailed characteristics of tissue spatial TME can provide valuable insight on disease progression and treatment response. Moreover, NACNet can be extended to explore the microenvironment reaction analysis with multiple subgraphs in future. 

The graph node awareness mechanism can enhance the efficacy of the graph-based WSI analysis. Note the NACNet-LIS model leverages node information, local textures, and SNA graph features, significantly improving the WSI-derived tissue graph representation and contextual understanding. With node features and the contextual node relationships, the NACNet-LIS with intensive graph node awareness allows for a comprehensive understanding of the tissue TME graph structure and properties and achieves a more nuanced understanding of the structure and properties of the graph. Such node awareness mechanism is particularly important in histopathology analysis, where subtle differences in tissue structure can have significant implications for diagnosis and treatment planning. Through ablation experiments, we have demonstrated the necessity of a comprehensive graph node characterization. 

Our work presents several avenues for improvement.  First, the reliance on detailed image annotations for histology prediction makes the deep learning model labor-intensive. Second, the clustering of discrete points with different labels to form graph nodes presents challenges. To improve node formation, we used a clinically-weighted clustering strategy that considers both the physical size differences and clinical significance of 12 histological categories in breast cancer. By assigning lower thresholds ($\eta$) to critical categories like tumor and necrosis, we ensured these features are prominently represented within the TME graph. For categories of lesser clinical importance, we used a higher threshold ($\eta$), reducing unnecessary segmentation while still maintaining relevant context. This approach not only minimizes misclassification impact but also enhances the model's ability to capture breast cancer’s pathological characteristics, supporting greater generalizability and clinical interpretability. Future experiments will further validate the effectiveness of this adjusted strategy, ensuring that the resulting TME graph structure holds higher biological significance and diagnostic value in breast cancer applications. Finally, increasing the sample size in future studies will be crucial to further validate the generalizability of our findings.

\section{Conclusion}
We present a transformer-based GCN model, NACNet, that predicts the NAC treatment response for TNBCs using WSIs of patient biopsy tissues. The model integrates local tissue histology information, tissue textures, and SNA features derived from the spatial TME graph to enhance NAC response prediction. The spatial attention mechanism and SNA features direct transformer GCN attention to tissue regions of prognostic significance, incorporating human prior knowledge. When applied to a TNBC patient dataset, our system achieves 90\% prediction accuracy. In addition, we demonstrate the efficacy of an integrative use of node content-awareness features, SNA information, and the GIN spatial attention mechanism on prediction enhancement. 
Our study also suggests the spatial TME histology patterns carry prognostic values for NAC response prediction for TNBC patients. Our graph-based analysis of histological variables for TME characterization and NAC response prediction is generic and can be tailored to enable treatment response prediction and personalized treatment planning for other cancer types.

\section*{Conflict of Interest Statement}
The authors declare that they have no known competing financial interests or personal relationships that could have appeared to influence the work reported in this paper.

\section*{Acknowledgments}
We thank Dr. Ritu Aneja from School of Health Professions at University of Alabama at Birmingham and Dr. Emiel A.M. Janssen from Department of Pathology at Stavanger University Hospital for providing the WSIs and WSI annotations. We would like to thank Timothy B. Fisher from Georgia State University for his valuable assistance in data preprocessing and performance evaluation of machine learning models for NAC treatment response prediction.

\section*{Funding}
This research was supported by Georgia State University Molecular Basis of Disease Doctoral Fellowship awarded to QL, and the Frady Whipple Endowment Professorship to YJ. 

\section{Supplement}
Supplementary Tables S1–2.

\bibliographystyle{cas-model2-names}

\bibliography{Nacnet}

\newpage

\end{document}